%% file: main.tex
\documentclass{article}

\usepackage[preprint]{corl_2026} % Use this for the initial submission.

\usepackage{amsmath}
\usepackage{amssymb}
\usepackage{amsthm}
\usepackage{enumitem}
\usepackage{microtype}
\input{sections/preamble}

\title{Decoding Task Progress from VLA Representations}

\author{
  Atiksh Bhardwaj\textsuperscript{*} \And
  Edward Weiyi Duan\textsuperscript{*} \And
  Pritwish Dan \And
  Wei-Chiu Ma \And Preston Culbertson \\ Cornell University
}

\begin{document}
\maketitle
\footnotetext[1]{Equal contribution. Correspondence to Atiksh \texttt{<ab2635@cornell.edu>}.}
\renewcommand{\thefootnote}{\arabic{footnote}}
\setcounter{footnote}{0}
%===============================================================================
\vspace{-20pt}
\begin{abstract}
    \input{sections/abstract}
\end{abstract}
\vspace{-6pt}
\keywords{Vision-Language-Action Models, Interpretability, Task Progress, OOD Detection} 

%===============================================================================
\section{Introduction}
\vspace{-2pt}
\label{sec:introduction}
\input{sections/introduction}
%===============================================================================
\vspace{-8pt}
\section{Related Work}
\vspace{-4pt}
\label{sec:relatedworks}
\input{sections/relatedwork}

%===============================================================================

%===============================================================================
\vspace{-4pt}
\section{Methods}
\vspace{-2pt}
\label{sec:methods}
\input{sections/methods}

%===============================================================================
\vspace{-4pt}
\section{Experiments and Results}
\vspace{-4pt}
\label{sec:results}
\input{sections/results}
\vspace{-7pt}
\section{Discussion and Limitations}
\vspace{-5pt}
\label{sec:discussion}
\input{sections/discussion}

%===============================================================================

\clearpage
% The acknowledgments are automatically included only in the final and preprint versions of the paper.
% \acknowledgments{If a paper is accepted, the final camera-ready version will (and probably should) include acknowledgments. All acknowledgments go at the end of the paper, including thanks to reviewers who gave useful comments, to colleagues who contributed to the ideas, and to funding agencies and corporate sponsors that provided financial support.}

%===============================================================================

\bibliography{references}  % .bib
\newpage
\renewcommand{\appendixpagename}{Appendix}
\appendixpage
\appendix

\section{Experimental Design}

\input{sections/expdesign}

\section{Additional Experiments}

\input{sections/moreexp}

\end{document}

%% file: sections/preamble.tex
\usepackage[table,dvipsnames]{xcolor}
\usepackage{bm}
\usepackage{graphicx}
\usepackage{float}
\usepackage{amsmath}
\usepackage{amssymb}
\usepackage{amsthm}
\usepackage{enumitem}
\usepackage{booktabs}
\usepackage{subcaption}
\usepackage{titlesec}
\usepackage{appendix}
\usepackage{algorithm}
\usepackage{algpseudocode}

\theoremstyle{definition}

\titlespacing*{\section}{0pt}{1.0ex plus 0.3ex minus 0.2ex}{0.6ex plus 0.2ex}
\titlespacing*{\subsection}{0pt}{0.1ex plus 0.1ex minus 0.1ex}{0.1ex plus 0.1ex}
\titlespacing*{\subsubsection}{0pt}{0.1ex plus 0.1ex minus 0.1ex}{0.1ex plus 0.1ex}

\renewcommand{\thefootnote}{\fnsymbol{footnote}}

%% file: sections/abstract.tex
Vision-language-action models (VLAs) are moving rapidly towards deployment as general-purpose manipulation policies, but we currently lack basic tools for understanding what these models represent internally or for monitoring them at runtime. Leveraging ideas from mechanistic interpretability, we probe the residual stream of $\pi_{0.5}$ and find that task progress, the normalized time remaining in a trajectory, is linearly readable from the activations. We find that this signal is present in the pretrained PaliGemma backbone prior to training on any robot-specific data. A single linear probe generalizes to unseen tasks and varies under language counterfactuals when trained on multi-prompt data, but does not enable meaningful steering of the policy. These properties make the signal directly useful for instrumenting deployed VLAs. We use the probe as a simple label-free OOD detector, which detects stalled task progress, and find it competitive with state-of-the-art methods. Our results suggest that VLAs have rich, linearly readable internal representations of semantic quantities like task progress, and that learning to read these signals offers a lightweight, interpretable path toward monitoring deployed visuomotor policies.

%% file: sections/introduction.tex
% Vision-Language-Action (VLA) models, which map raw visual observations and natural language instructions directly to continuous control, have emerged as a leading approach to scaling robot learning beyond narrow, task-specific systems. Recent models such as $\pi_{0.5}$~\citep{intelligence2025pi05}, OpenVLA~\citep{kim2025openvla}, and RT-2~\citep{zitkovich2023rt2} leverage large-scale cross-embodiment pretraining~\citep{oneill2024openx} to inherit the broad semantic and visual knowledge of their underlying vision-language backbones. In practice, however, deploying these models on new robots, environments, or task suites typically requires task-specific fine-tuning~\citep{kim2025openvla, intelligence2025pi05}. This fine-tuning has been shown to degrade the very generalization that pretraining was meant to provide: fine-tuned models lose language-following capability~\citep{driess2025ki}, overfit to scene memorization~\citep{swann2026drvla, zhou2025liberopro} \preston{What does this phrase mean?}, and fail under perceptual perturbations they would otherwise handle~\citep{kachaev2025dontblind} \preston{Similar question - can we be more specific about the finding?}. These failures motivate runtime monitoring: detecting at inference time when a policy is drifting out of distribution or heading toward task failure, so that execution can be halted, deferred, or handed off~\citep{gu2025safemultitaskfailuredetection, xu2025faildetect, romer2025fiper}.

Vision-Language-Action models (VLAs) promise a path to generalizable robot manipulation. By coupling a pretrained vision-language model (VLM) backbone with an action decoding head, they inherit rich semantic and visual priors which can support open-vocabulary task specification across diverse environments and embodiments. In practice, however, deploying VLAs on new robots or tasks requires task-specific fine-tuning~\citep{kim2024openvla, intelligence2025pi05}, which systematically degrades the strong generalization that motivated the use of a VLM backbone. After fine-tuning, VLAs have been shown to lose language sensitivity~\citep{driess2025knowledge}, to become brittle to perceptual perturbations~\citep{kachaev2025dont}, and can collapse to near-zero success when training-scene assumptions are perturbed~\citep{zhou2025libero-pro}. While these failures are common and well-documented, we fail to leverage pre-existing tools to understand when or why a VLA will fail, or to monitor their internal state at runtime.

Interpretability tools developed for vision and language models~\citep{bau2017network, templeton2024scaling, radford2021learning, marks2023geometry, park2024linear} transfer naturally to VLAs,
whose pretrained VLM backbones~\citep{beyer2024paligemma} share the same substrate. Taken together, these results establish VLA internals as a tractable target for interpretability. Recent work has begun to confirm this empirically~\citep{buurmeijer2026observing, haon2025mechanistic,lu2025probing,molinari2025emergent,swann2026sparse}. These efforts establish that VLA representations are readable, but they characterize what the policy knows about the world, not where the policy stands relative to its task. The latter is a distinct class of feature, scalar, task-conditional, and directly tied to behavior, and it remains unprobed.

We focus on one such signal: \emph{task progress}, the normalized fraction of a trajectory remaining until completion. Progress is a foundational quantity in sequential decision-making; it serves as a natural goal-conditioned value signal, monotonically tracking the policy's distance to task completion~\citep{ma2023vip}. A substantial body of work in robot learning has trained external vision-language models to estimate progress from observations, deploying these estimators as reward signals, planning heuristics, and success detectors~\citep{ma2023liv, ma2025vision}. But VLAs already process the same visual and language inputs these external estimators consume, and produce actions that implicitly depend on progress through the task. We hypothesize that progress is already encoded in the policy's own internal representations, recoverable without training a separate model at all.

% \vspace{-2mm}
% \begin{enumerate}[leftmargin=*]
% \itemsep0em 
% \item We identify task progress as a linearly decodable, task-shared feature in VLA representations, distinct from prior probed feature classes such as low-level dynamics, discrete symbolic predicates, and unsupervised semantic features.
% \item We demonstrate that the progress probe serves as an effective runtime out-of-distribution detector under perceptual perturbation, with cross-task generalization to held-out tasks, complementing prior runtime failure detectors that learn dedicated failure classifiers on VLA features~\citep{gu2025safemultitaskfailuredetection}. (UPDATE WITH SAFE RESULTS?)
% \item Controllability? Steering, if we can get it to work?
% \item We find that the progress feature is dominantly visual despite language-conditioned training, with a contrastive ablation confirming that language-conditioned progress is not linearly recoverable from the policy's representations. (MAYBE)
% \end{enumerate}

In this work, we apply linear probing to $\pi_{0.5}$~\citep{intelligence2025pi05} to study what its representations encode about task progress. We confirm the hypothesis: progress is reliably decodable from intermediate embeddings, and the resulting probe supports several downstream uses. We use it as a runtime out-of-distribution detector competitive with supervised baselines~\citep{gu2025safe}, as a diagnostic for language-grounding degradation under fine-tuning~\citep{driess2025knowledge, kachaev2025dont}, and as a tool for studying the gap between feature observability and controllability~\citep{buurmeijer2026observing}, finding that task progress is observable but not linearly controllable. Our contributions are as follows:

\begin{enumerate}[nosep,leftmargin=*]
\itemsep0em
\item We present definitions for \textbf{weak decodability}, \textbf{strong decodability}, and \textbf{steerability} for generative models, separating the existence of a linearly correlated signal from its dependence on the correct inputs and its causal effect on outputs.
\item We show that \textbf{task progress} is both weakly and strongly decodable from $\pi_{0.5}$'s residual stream but not steerable: a single linear probe recovers progress, varies correctly under language counterfactuals, but fails to enable model steering via injection.
\item We show that the progress probe, when used at runtime as a label-free \textbf{OOD detector}, is competitive with supervised baselines, with strong cross-task and cross-perturbation generalization.
\end{enumerate}

\begin{figure}[tb]
    \centering
    \includegraphics[width=\textwidth]{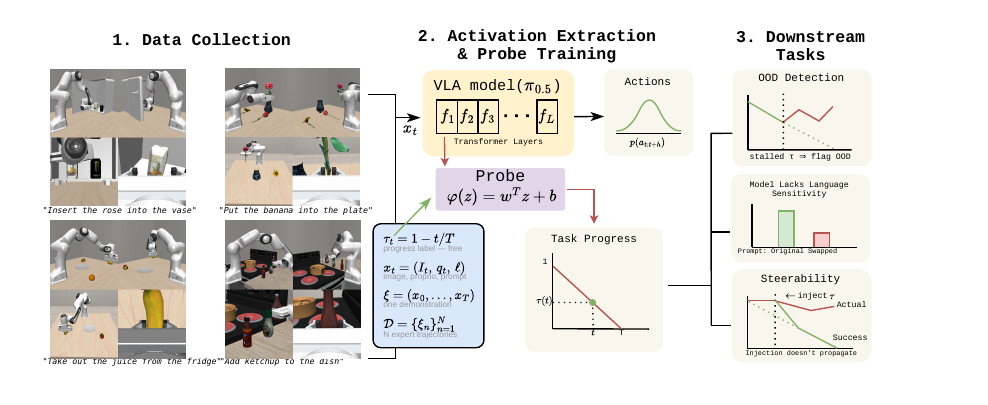}
    \caption{\textbf{Task progress is linearly readable from $\pi_{0.5}$'s internal activations.} A linear probe $\varphi(\bm{z}^0) = \bm{w}^\top \bm{z}^0 + b$ on residual-stream activations $\bm{z}^0$ from the VLM backbone recovers task progress, the normalized fraction of the trajectory remaining, and supports three downstream analyses: OOD detection, a language-grounding diagnostic, and a steerability study.}
\vspace{-8pt}
    \label{fig:figure_1}
\end{figure}

%% file: sections/relatedwork.tex
\subsection{Interpretability for Large Models}

% With the advent of VLAs such as $\pi_{0.x}$ \cite{black2024pi0, intelligence2025pi05}, OpenVLA \cite{kim2025openvla}, and RT-2 \cite{zitkovich2023rt2}, there has been a growing interest in understanding the inner workings of these models. This has led to a recent push for developing interpretability techniques for VLAs, following recent successes in interpretability for large language models (LLMs). 

A growing body of work in neural network interpretability has built on the \emph{linear representation hypothesis}, ~\citep{park2024linear, mikolov2013linguistic}, which holds that high-level semantic concepts in large pretrained models tend to be linearly encoded in the activation spaces. Evidence for this hypothesis has accumulated for both vision 
~\citep{radford2021learning} and language models ~\citep{marks2023geometry, park2024linear}. Notably, transformers trained to predict Othello moves develop linearly decodable representations of the board state \cite{li2023emergent, nanda2023emergent}, suggesting that models trained purely for imitation can develop structured world models. Beyond analysis, this linear structure supports training-free steering via contrastive concept directions \cite{zou2023representation} and detection of behaviors not visible in model outputs \cite{hubinger2024sleeper, macdiarmid2024simple}.

% Taken together, these results indicate that large model representations contain interpretable, geometrically structured encodings of semantic concepts that are useful for monitoring and steering, and provide a toolkit we apply here to VLAs.

Several recent works apply interpretability methods to VLAs directly. \citet{haon2025mechanistic} use mechanistic techniques to identify semantic directions in VLA activations, and \citet{swann2026sparse} use sparse autoencoders to find ``monosemantic'' directions correlated with motion primitives with both exploring these for steering. \citet{buurmeijer2026observing} decode and steer actions and physical states with linear probes, finding physical states cleanly decodable from internal activations, and high-level logical states (e.g., one object resting on another) are likewise decodable from OpenVLA~\citep{lu2025probing}. Our work joins this line but differs in focus: rather than cataloguing which signals are present in activations or steering the model directly, we target one specific signal, task progress, for runtime monitoring.

\subsection{Out of Distribution Detection}
\vspace{-2pt}
A core safety concern for VLAs is that they often fail to generalize to out-of-distribution (OOD) states including visual perturbations, unseen task variants, or novel environments, which can produce silent failures during deployment. Prior work on OOD detection in deep models falls into two broad families. \textit{Density and distance-based} methods, such as Mahalanobis-distance scoring~\citep{lee2018simple} and its PCA-projected variants, energy-based detection~\citep{liu2020energy}, and variational autoencoders, fit a model of the in-distribution feature distribution and flag low-density points. \textit{Supervised} approaches train detectors directly on labeled in-distribution and out-of-distribution data: SAFE~\citep{gu2025safe} adapts this to VLAs by training MLP/LSTM heads on the model's internal embeddings. Conformal-prediction methods~\citep{farid2022failure} provide statistical failure guarantees, and language-model runtime monitors~\citep{sinha2024real} flag anomalies by reasoning over the policy's observations. 

Both families share limitations for VLAs: density-based detectors memorize the training-task distribution, while supervised detectors require OOD labels and transfer poorly across perturbations. We instead use \emph{task progress}, a feature intrinsic to the policy, as a calibrated OOD signal requiring no OOD supervision and tying failure detection to the model's own representation of task completion.

% A core issue with VLAs is that they often fail to generalize primarily to out-of-distribution (OOD) states within tasks and with new environments or tasks, which can lead to safety concerns in real-world applications. Several methods have previously proposed to address this issue, including uncertainty estimation, anomaly detection, and domain adaptation. Recently, a push has been to use VLA embeddings for failure detection and OOD detection, with methods such as SAFE [SAFE Paper] showing promising results in detecting OOD states by leveraging the internal representations of the VLA. 

% As such, we choose to evaluate the progress feature in the VLA model for its potential use in OOD detection, as it encodes the progress towards task completion and may be sensitive to changes in the environment or task that could indicate OOD states. We will compare the 

%% file: sections/methods.tex
\subsection{Vision-Language Action Models and Embedding Extraction}
\label{subsec:embedding-extraction}
We define the transformer architecture commonly found in VLAs including our targeted VLA model of $\pi_{0.5}$~\citep{intelligence2025pi05} and its vision-language backbone of PaliGemma~\citep{beyer2024paligemma}. Given an input observation $\bm{x} = (\bm{I}, \bm{q}, \ell)$ which consists of an image $\bm{I}$, proprioceptive data $\bm{q}$, and a language prompt $\ell$, the VLA model processes this sequence through multiple layers of transformer blocks, producing a sequence of intermediate embeddings at each layer. 

As the input $\bm{x}$ is passed through a sequence of $L$ layers, it turns into an embedding $\bm{z}^i \in \mathbb{R}^d$ for layer $i = 0, \dots, L$. Each embedding is computed sequentially via a residual function $\bm{f}^i: \mathbb{R}^d \to \mathbb{R}^d$,\begin{equation*}
    \bm{z}^{i + 1} = \bm{z}^i + \bm{f}^{i} (\bm{z}^i), \quad i = 0, \dots, L-1.
\end{equation*}
We extract the embeddings from the targeted VLA by computing the intermediate embeddings $\bm{z}^i_t$ for each observation $\bm{x}_t$ in the post-training dataset $\mathcal{D} = \{(\bm{x}_t, \bm{a}_t)\}_{i=0}^{N}$ for $t = 0, \dots, T$ and $N$ trajectories.  For each input $\bm{x}_t$, we cache the intermediate embeddings $(\bm{z}_t^0, \bm{z}_t^1, \dots, \bm{z}_t^L)$, and the corresponding labels for the target feature $\zeta$ for each time step $t$, creating a dataset for probe training: $\mathcal{D}_\text{probe} = \{(\bm{z}_t^0, \bm{z}_t^1, \dots, \bm{z}_t^L, \zeta_t)\}_{i=0}^N$. Additionally, each embedding $\bm{z}_t^i$ will have a shape $d \times n$, where $d$ is the size of the embedding and $n$ is the number of tokens, where we mean-pool across the tokens. 

    \subsection{Linear Probes for Feature Decodability and Steerability}
    Claims that a network ``represents'' a feature vary in strength, from ``a correlated signal can be extracted'' to ``editing the signal steers the model'' \cite{hewitt2019designing}. We formalize three notions along this spectrum.

    Consider a layer $i$ and a scalar feature $\zeta \in \mathbb{R}$ of the input. Let $\varphi(\bm{z}^i) = \bm{w}^\top \bm{z}^i + b$ be a linear probe which seeks to estimate $\zeta$ from $\bm{z}^i$, fit on $\mathcal{D}_\text{train} = \{(\bm{z}^i_k, \zeta_k)\}_{k=1}^K$. As a running example, consider a VLA performing tabletop manipulation, where $\zeta(\bm{x})$ is the height of the target object referenced in the language prompt $\ell$ (e.g., \emph{``pick up the red mug''}). For any dataset $\mathcal{D} = \{(\bm{z}^i_k, \zeta_k)\}_{k=1}^K$, we define the probe's risk as
    \begin{equation*}
        R(\varphi; \mathcal{D}) = \frac{1}{K}\sum_{k=1}^K \mathcal{L}\bigl(\varphi(\bm{z}^i_k); \zeta_k\bigr),
    \end{equation*}
    where $\mathcal{L}$ is a loss appropriate for the feature type (e.g., squared error for continuous $\zeta$, cross-entropy for discrete $\zeta$). We can use this risk to discuss different notions of ``representation.''
    
    \paragraph{Weak decodability.} Evaluating $R(\varphi;\,\mathcal{D}_{\text{val}})$ on $\mathcal{D}_{\text{val}}$ drawn from the training distribution tests \emph{weak decodability}: whether a linear signal correlated with $\zeta$ can be extracted from in-distribution activations. However, low risk does not rule out spurious correlations (e.g., the probe always predicting the red mug's coordinate if the mug was the only target object in $\mathcal{D}_\text{train})$.
    
    \paragraph{Strong decodability.} To address this, let $\mathcal{T}: \mathcal{X} \to \mathcal{X}$ denote a counterfactual transformation that perturbs the input factors on which $\zeta$ depends. Evaluating $R(\varphi;\,\mathcal{D}_{\mathcal{T}})$, where $\mathcal{D}_{\mathcal{T}} = \{(\bm{z}^i(\mathcal{T}(\bm{x}_k)),\, \zeta(\mathcal{T}(\bm{x}_k)))\}_{k=1}^K$, tests \emph{strong decodability}: whether the same probe, without retraining, generalizes to counterfactually transformed inputs. For instance if $\mathcal{T}_\text{swap}$ replaces the language prompt $\ell$ with a different instance $\ell^\prime$ (e.g., \emph{``pick up the blue bowl''}), low risk on $\mathcal{D}_\mathcal{T}$ indicates that the probe would respond to this switch, which rules out the ``red mug'' shortcut.
    
    \paragraph{Steerability.} However, strong decodability still does not establish \emph{steerability}: whether the
    model uses this signal for decision-making.  To test this, we
    intervene directly on the activations.  Let $\pi(\bm{z}^i)$ denote
    the distribution of actions produced by the model when the
    activations at layer~$i$ are $\bm{z}^i$.  For a counterfactual
    input~$\mathcal{T}(\bm{x})$, consider the steered activation
    \begin{equation} \label{eq:steer}
      \tilde{\bm{z}}^i(\bm{x})
      \;=\;\bm{z}^i(\bm{x})
           + \frac{\zeta(\mathcal{T}(\bm{x}))-\zeta(\bm{x})}
                  {\|\bm{w}\|^2}\,\bm{w},
    \end{equation}

    which shifts the activation along $\bm{w}$ so that the probe approximately reads
    out the correct counterfactual value,
    $\varphi(\tilde{\bm{z}}^i)\approx\zeta(\mathcal{T}(\bm{x}))$. Let $\tilde{\pi}_k = \pi(\tilde{\bm{z}}^i(\bm{x}_k))$ denote the steered action distribution and $\pi^{\star}_k = \pi(\bm{z}^i(\mathcal{T}(\bm{x}_k)))$ for the true counterfactual.  The \emph{steerability risk} measures how well steering approximates the true counterfactual, and is defined as
    \begin{equation} \label{eq:steer_risk}
      R_S(\varphi;\,\mathcal{D},\,\mathcal{T})
      \;=\; \frac{1}{K}\sum_{k=1}^{K}
        d\bigl(\tilde{\pi}_k,\;\pi^{\star}_k\bigr),
    \end{equation}
    where $d$ is a divergence between action distributions.  In our example, low~$R_S$ under~$\mathcal{T}_{\text{lift}}$, which generates images where the red mug is at a higher $z$-coordinate, means that injecting a higher mug coordinate using $\varphi$ causes the model to generate action chunks as if the mug were at this location.

\subsection{The Progress Feature}

We define our targeted feature, \textbf{task progress} $\tau$, as the following: given a trajectory $\xi= (\textbf{x}_0, \textbf{x}_1, \dots, \textbf{x}_T)$ where each observation $\textbf{x}_t = (\textbf{I}_t, \textbf{q}_t, \ell)$ contains images, proprioception, and language prompt. We first define a timestep mapping $\phi_{\xi}: \mathcal{X} \to \{0, 1, \dots, T\}$ that returns the index of an observation within $\xi$ such that $\tau$ is the normalized time remaining:
\begin{equation*}
\phi_{\xi}(\textbf{x}_t) \;=\; t,
\qquad
\tau(\textbf{x}_t) \;=\; 1 - \frac{\phi_{\xi}(\textbf{x}_t)}{T} \;\in\; [0, 1]
\label{eq:progress}
\end{equation*}

Task progress is a useful signal for runtime monitoring because it satisfies several desirable properties: it decreases monotonically along expert trajectories (yielding predictable in-distribution behavior), can be easily labeled from offline data, readily flags out-of-distribution states (when progress stalls), and generalizes across tasks.

% Via Eq. \ref{eq:progress}, we find that the following properties emerge through our experiments:

% \begin{enumerate}[nosep,leftmargin=*]
%     \item \textbf{Monotonicity:} $\tau(\textbf{x}_t) \geq \tau(\textbf{x}_{t'})$ for all $t \leq t'$ along an in-distribution (ID) trajectory.
%     \item \textbf{Coherent OOD behavior:} Let $\hat{\tau} = \bm{\varphi}^i(\textbf{z}_t^i)$ denote the probe estimate of $\tau$ at time $t$. For a trajectory that enters an OOD state at timestep $t^* \leq T$, $\hat{\tau} - \tau(\textbf{x}_t) > 0$ for all $t > t^*$, i.e., the probe estimate stagnates above the in-distribution progress curve $1 - t/T$.
%     \item \textbf{Language sensitivity:} $\tau$ is conditioned on the language prompt $\ell$ used to train the probe; observations sharing $(\textbf{I}_t, \textbf{q}_t)$ but differing in $\ell$ yield different progress values.
% \end{enumerate}

\subsection{Out of Distribution (OOD) Detection via the Progress Feature}
\label{sub:oodprogress}

We find the progress signal a strong indicator for when the policy goes OOD enabling us to develop a method to flag OOD based on the progress feature decoded via our probes.

The detector takes as input a progress probe $\varphi$
and a threshold $\delta$. The expected task completion time $\mathbb{E}[T \mid \ell]$
is estimated once from in-distribution rollouts. At runtime, for each replan, we decode the predicted progress
$\tau_{\text{pred}} = \varphi(\bm{z}_t^i)$ using the probe. We then compute the residual
\begin{equation}
  V_\tau(t) = \tau_{\text{pred}} - \left(1 - \frac{t}{\mathbb{E}[T \mid \ell]} \right),
\end{equation} which measures the deviation between the predicted and actual progress. The rollout is flagged as out-of-distribution the first time
$V_\tau(t) > \delta$.

While simple, this procedure is able to detect stalled progress using the probe alone, which is both task-agnostic (assuming $\mathbb{E}[T \mid \ell]$ is known or easily estimated) and does not require OOD labels.

% \begin{enumerate}[nosep,leftmargin=*]
%     \item Build a per-task, per-prompt calibration of expected progress on the ID dataset.
%     \item Train a linear probe on the same data and denote the resulting observer as $\bm{\varphi}^i$.
%     \item At each model replan, compute $\bm{\varphi}^i(\bm{z}_t^i) = \tau_{\text{pred}}$.
%     \item Define $\bm{V}_{\tau} = \tau_{\text{pred}} - \mathbb{E}[\tau \mid t]$ as the deviation between the predicted $\tau$ and the expected $\tau$.
%     \item If $\bm{V}_{\tau} > 0$, the rollout has gone OOD: expected progress continues while $\tau_{\text{pred}}$ has stalled.
% \end{enumerate}

%% file: sections/results.tex
\subsection{Experimental Setup}
\label{sub:experimental}

Our experiments evaluate the progress feature along four axes: weak decodability, strong decodability, steerability, and OOD detection. The target model is  $\pi_{0.5}$, fine-tuned on VLABench~\citep{zhang2025vlabench}, which provides 10 manipulation tasks with semantic variation and 500 trajectories per task. We follow the procedure from Subsection \ref{subsec:embedding-extraction} to collect the embeddings and labels for $\tau$. 

% We mean-pool across the token axis to obtain a $d$-dimensional vector per embedding $\bm{z}_t^i$, and each observer $\bm{\varphi}^i$ is then trained on those embeddings resulting a $d$-dimensional vector for $\bm{\varphi}^i$.

\subsection{Where does the progress feature emerge from?}
\label{subsec:weak}

\begin{figure}[tb]
    \centering
    \vspace{-10pt}
    \includegraphics[width=0.9\textwidth]{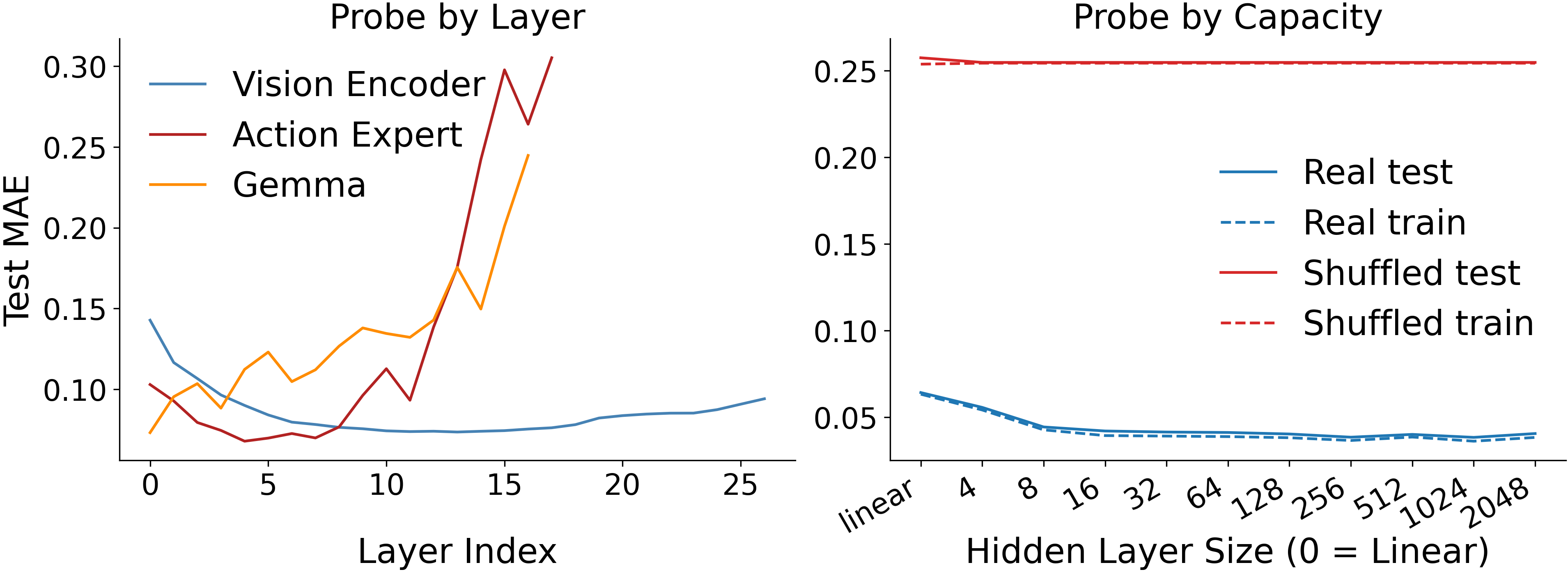}
    \caption{\textbf{Left:} Weak decodability of the progress feature across layers of fine-tuned $\pi_{0.5}$ (SigLIP, PaliGemma's Gemma, and the action expert). \textbf{Right:} Probe performance vs. capacity on the first Gemma layer, true labels versus shuffled labels}
    \label{fig:observability}
\end{figure}
\emph{Q1: Can we weakly decode the progress feature across the layers of the VLA model?} We train one linear probe per layer of $\pi_{0.5}$ under an $L_1$ loss (Fig.~\ref{fig:observability}, left), with the vision-encoder curve averaged across the front, side, and wrist camera embeddings. We find progress is weakly decodable (i.e., achieves a low in-distribution loss) in the early vision-backbone layers and the first few Gemma layers. Given these results, we probe layer 0 for weak decodability and OOD detection
(Sec.~\ref{subsec:ood}), and layer 10 for strong decodability
(Sec.~\ref{subsec:strong}). Because a sufficiently expressive probe can fit random targets \cite{hewitt2019designing}, we also train more expressive MLP probes on a control task, where all labels are shuffled. If a probe can memorize these random labels, then any  signals would be weakly decodable at this layer, and this test is not \emph{selective}. In Fig.~\ref{fig:observability} (right), we plot the gap between the true-label probe and the shuffled-label probe across a sweep of MLP capacities, finding a large gap even for high-capacity probes, indicating that task progress being weakly decodable is a meaningful property of the layer. 

% We train one linear probe per layer of the VLA model under an L1 loss and report performance in Fig.~\ref{fig:observability} (left); the vision-encoder curve is averaged across the front, side, and wrist camera embeddings. Progress is weakly decodable in the early layers of the vision backbone and the first few layers of Gemma, indicating that it emerges early in the processing pipeline. We use the first Gemma layer for the remainder of the experiments, as task progress should be conditional on language input and this is the first layer at which the language and vision modalities are fused. To rule out spurious correlations within the embeddings, we retrain the probes on the same embeddings with shuffled progress labels. We expect, that if we scale the hidden size in a single layer, multi-layer perceptron (MLP), the MLP fails to find patterns on the shuffled set, especially at lower hidden sizes. In contrast, we should see the MLP easily learn the $\tau$ feature from the embeddings. Fig.~\ref{fig:observability} (right) shows that the true-label probe substantially outperforms the shuffled-label probe on both train and test MAE across probe capacities confirming that the probe is recovering a genuine progress signal rather than noise.

% \textbf{Q2:} How observable is the progress feature prior to fine-tuning on the VLABench dataset?

\begin{figure}[b]
    \centering
    \includegraphics[width=0.9\textwidth]{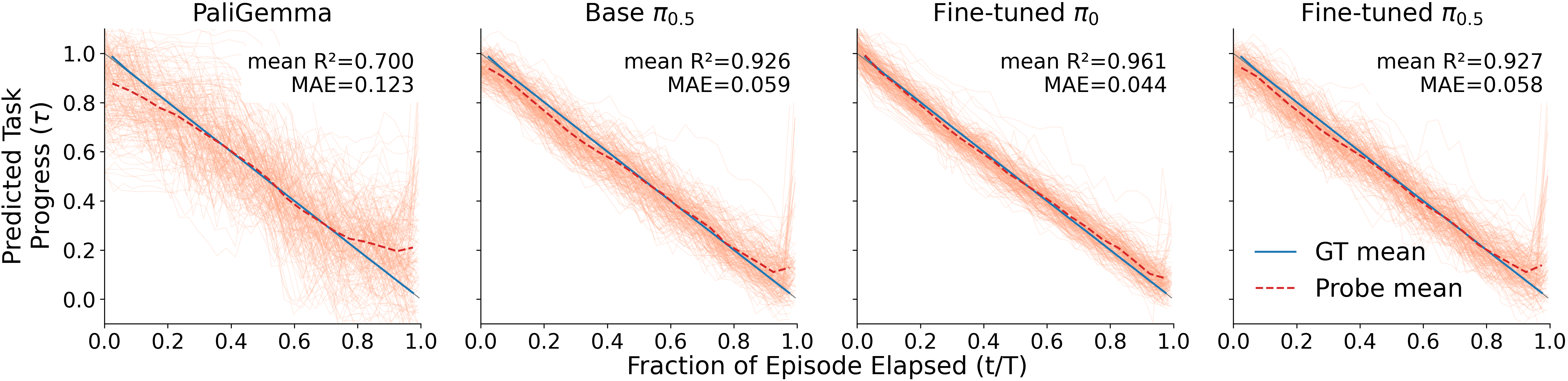}
    \caption{Weak decodability of the progress feature across four models, probed at successive stages of training: \textbf{PaliGemma} (vision-language backbone only), \textbf{base $\pi_{0.5}$} (after large-scale robot pre-training, before fine-tuning), and \textbf{VLABench fine-tuned $\pi_0$} (a distinct VLA architecture), and \textbf{VLABench fine-tuned $\pi_{0.5}$} (our target model).}
    \label{fig:model_comparison}
    \vspace{0pt}
\end{figure}

\emph{Q2: Is the progress feature weakly decodable prior to fine-tuning and across architectures?} 
We compare probe performance across four models chosen to isolate which stage of training establishes the progress feature: the PaliGemma vision-language backbone, base $\pi_{0.5}$ prior to fine-tuning, VLABench fine-tuned $\pi_0$, and VLABench fine-tuned $\pi_{0.5}$ (Fig.~\ref{fig:model_comparison}). Probes are trained on 400 episodes (100 per task) and evaluated on 50 held-out episodes per task. Fig.~\ref{fig:model_comparison} shows that the feature is observable in all four models, but the three $\pi_{0.x}$ variants substantially outperform PaliGemma and perform comparably to one another. This indicates that the progress feature is established by large-scale robotic pre-training and not by fine-tuning.

\subsection{How does the progress feature react to language perturbations?}
\label{subsec:strong}
\vspace{-4pt}
We test for strong decodability with respect to a language-perturbation counterfactual $\mathcal{T}_\text{lang}$, which swaps the target noun in the instruction (e.g., \emph{``put the pear onto the plate''} $\to$ \emph{``put the apple onto the plate''}) while leaving the visual observation and proprioception unchanged. For each demonstration from the expert dataset, we run several ``shadow'' forward passes under unique counterfactuals $\mathcal{T}_\text{lang}$ and cache the Gemma activations for each layer.

\begin{figure}[tb]
  \centering
  \includegraphics[width=0.9\textwidth]{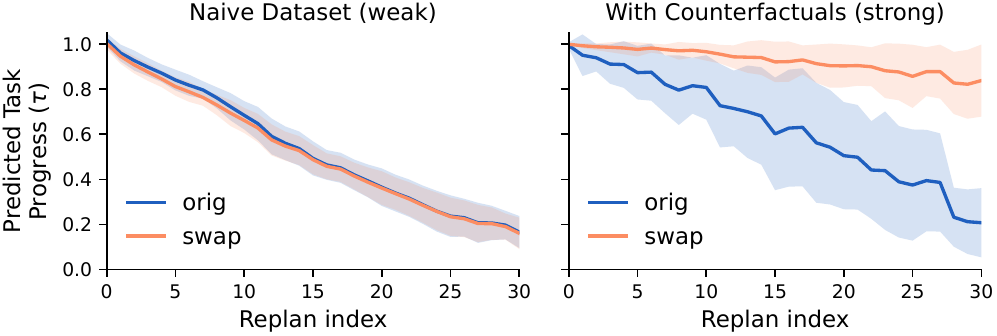}
  \caption{\textbf{Counterfactual training yields strong decodability at a cost.}
  Progress traces under the original instruction (blue) and $\mathcal{T}_{\text{lang}}$ noun-swap (orange). \textbf{Left:} naive layer-0 probe. \textbf{Right:} contrastive probe at layer 10, trained with $\mathcal{T}_{\text{lang}}$ counterfactuals labeled $\tau = 1.0$. Bands: 95\% CI over 75 rollouts.}
  \vspace{-6pt}
  \label{fig:strong-decodability}
\end{figure}

\emph{Q1: Does a naively trained probe respond to language perturbations? }
The layer-0 probe from ~\ref{subsec:weak} decodes task progress $(R^2 = 0.95)$, but its output under $\mathcal{T}_\text{lang}$ is nearly identical to its output under the original instruction (Fig.~\ref{fig:strong-decodability}, left). Likely, this means the probe relies solely on vision and ignores the language input $\ell$. Task progress is not strongly decodable using this probe.

\emph{Q2: Does counterfactual training yield strong decodability?}
We augment $\mathcal{D}_{probe}$ with $\mathcal{T}_\text{lang}$ activations for
counterfactual prompts, labeled with $\tau = 1.0$, and retrain at layer 10 with the contrastive objective of App.~\ref{app:contrastive}. The label is a proxy: the true progress on a swapped task is hard to define and impossible to collect without separate counterfactual rollouts, so we approximate it by treating shadow passes as making zero progress on the instructed task. ID $R^2$ falls to $0.33$, reflecting both the noisy proxy and the lower progress ceiling at layer 10 (App.~\ref{app:contrastive}).
The tradeoff is favorable: the probe now pins near $\tau = 1.0$ under
$\mathcal{T}_\text{lang}$ while tracking progress normally under the original
instruction (Fig.~\ref{fig:strong-decodability}, right), recovering the language
sensitivity that the naive probe lacked.

\vspace{-4pt}

\subsection{Is the progress feature steerable?}
\label{subsec:steerable}

\begin{figure}[b]
    \centering
    \vspace{-12pt}
    \includegraphics[width=0.9\textwidth]{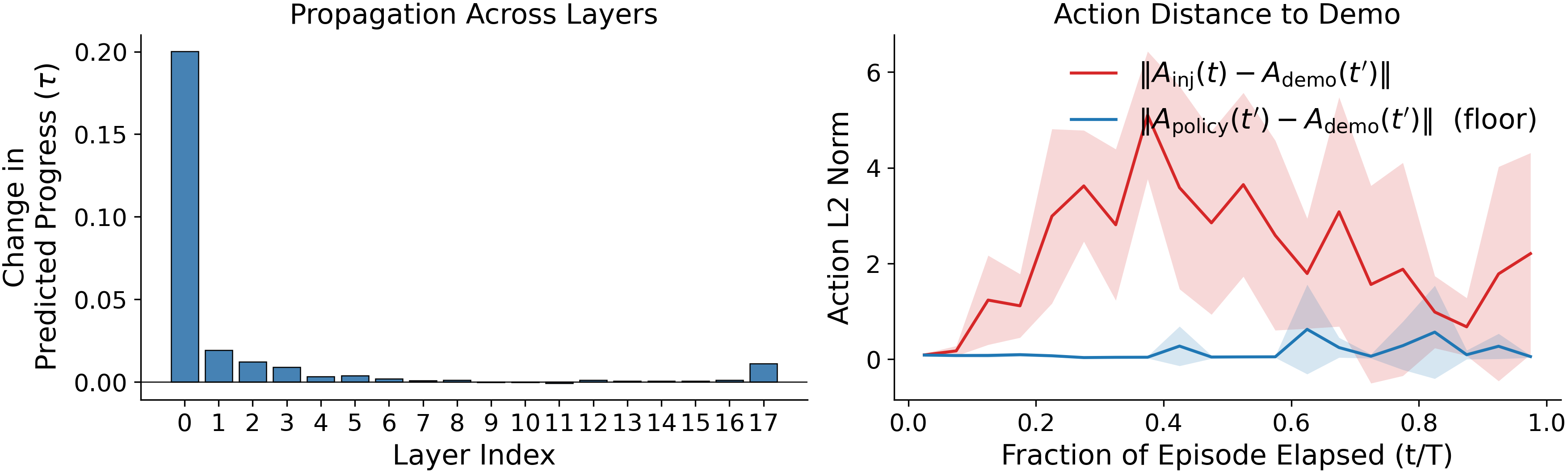}
    \caption{\textbf{Left:} Predicted progress across Gemma layers $i = 0, \dots, 17$ after injecting $\tau_{\text{boost}} = 0.2$ at layer 0 via $\bm{\varphi}^0$, read at each subsequent layer by $\bm{\varphi}^i$. \textbf{Right:} $\|A_{\text{inj}} - A_{\text{GT}}\|_2$ compared against the non-injected baseline $\|A_{\text{policy}} - A_{\text{GT}}\|_2$, with 95\% confidence interval (CI) over 5 rollouts run 5 times each. A successful injection would bring the injected curve close to the policy baseline.}
    \vspace{-8pt}
    \label{fig:injection}
\end{figure}
% \begin{figure}[tb]
%     \centering
%     \includegraphics[width=1.0\textwidth]{figures/injection_plot.png}
%     \caption{\textbf{Left:} Predicted progress across Gemma layers $i = 0, \dots, 17$ after injecting a 20\% forward progress signal at layer 0, using the direction recovered by a linear probe trained on layer 0 embeddings. Progress is read at each subsequent layer by a probe trained on that layer's embeddings. \textbf{Right:} L2 distance between the injected action $A_{\text{inj}}$ and the ground-truth future action $A_{\text{GT}}$, compared against the L2 distance between the non-injected policy action $A_{\text{policy}}$ and the same $A_{\text{GT}}$ both with 95\% confidence interval error bounds. A successful injection would bring the injected curve close to the policy baseline.}
%     \label{fig:injection}
% \end{figure}

We aim to steer the policy's output based on $\tau$. We define $\mathcal{T}_{\text{boost}} : \mathcal{X} \to \mathcal{X}$ by $\mathcal{T}_{\text{boost}}(\bm{x}_t) = \bm{x}_{t^\star}$, where $\frac{t^\star - t}{T} = \tau_{\text{boost}}$ injecting $\tau_{\text{boost}}$ forward progress; here $t$ is the current timestep, $t^\star$ the future timestep, and $T$ the total number of timesteps. We train one linear probe per Gemma layer, yielding 18 observers $\bm{\varphi}^i$ for $i = 0, \dots, 17$. We then use $\bm{\varphi}^0$ as the injection direction on layer 0 via Eq.~\ref{eq:steer} with $\tau_{\text{boost}} = 0.2$, and read the resulting progress at each subsequent layer using $\bm{\varphi}^j$ for $j = 1, \dots, 17$. Fig.~\ref{fig:injection} (left) shows that the predicted progress is unchanged beyond layer 0, indicating that the layer-0 injection does not boost progress across layers. To test whether the injection steers the policy toward future behavior, we decode an injected action $A_{\text{inj}}$ from each perturbed embedding across five random demonstrations (five injections each) and compare $\|A_{\text{inj}} - A_{\text{GT}}\|_2$ to the non-injected baseline $\|A_{\text{policy}} - A_{\text{GT}}\|_2$, where $A_{\text{GT}}$ is the corresponding future action from the demonstration, estimating Eq.~\ref{eq:steer_risk}. A successful injection would draw the two distances together, but Fig.~\ref{fig:injection} (right) shows the injected actions remain at a large, non-zero distance while the policy baseline lies near zero. We conclude that the progress feature $\tau$ is not steerable.

\vspace{-3pt}

\subsection{Does the progress feature serve as a reliable Out of Distribution detector?}
\label{subsec:ood}
\vspace{-4pt}
The progress feature provides a strong signal for identifying states within the policy's training distribution. When the policy enters OOD states, it can no longer advance the task, so its progress estimate should stagnate, yielding both a qualitative and a quantitative signal for OOD detection.

We generate OOD states via four naturally-motivated image augmentations of ID states: (i) \textbf{gaussian noise} (seen in low-light capture); (ii) \textbf{color shift} (caused by changes in lighting); (iii) \textbf{blur} (caused by camera shake during manipulation); and (iv) \textbf{occlusion} (from obstacles in the scene).

% \begin{enumerate}[nosep,leftmargin=*]
%  \item \textbf{Gaussian Noise:} Seen in low-light capture
%  \item \textbf{Color Shift:} Caused by changes in lighting via time or light sources
%  \item \textbf{Blur:} Caused by camera shake during manipulation
%  \item \textbf{Occlusion:} From obstacles or moving objects in the scene
% \end{enumerate}

% \begin{enumerate}[nosep,leftmargin=*]
%  \item \textbf{Gaussian Noise:} seen in low-light capture; additive per-pixel noise $\mathcal{N}(0, \sigma \cdot 255)$ with $\sigma = 0.1$.
%  \item \textbf{Color Shift:} from changes in lighting via time or light sources; uniformly applying random HSV shift with hue $\in [-\sigma \cdot 180, \sigma \cdot 180]$ degrees and saturation/value $\in [-\sigma \cdot 255, \sigma \cdot 255]$ at $\sigma = 0.3$.
%  \item \textbf{Blur:} from camera shake during manipulation; gaussian blur with kernel size $k = 15$ and standard deviation $\sigma = 0.3$.
%  \item \textbf{Occlusion:} from obstacles or moving objects in the scene; three random occlusions per image, each of size $(\sigma H, \sigma W)$ with $\sigma = 0.2$.
% \end{enumerate}

For each trajectory, we introduce the perturbation starting at one of three points, $k \in \{0.25, 0.5, 0.75\}$ of the trajectory length, and let the policy act on the perturbed input, collecting the rollout online until the episode ends. We sample 300 ID trajectories across six tasks (50 per task) and, for each noise mode and injection point, generate 50 OOD trajectories per task. We then evaluate robustness to (i) unseen tasks and (ii) unseen OOD conditions.

\begin{figure}[tb]
    \centering
    \includegraphics[width=1.0\textwidth]{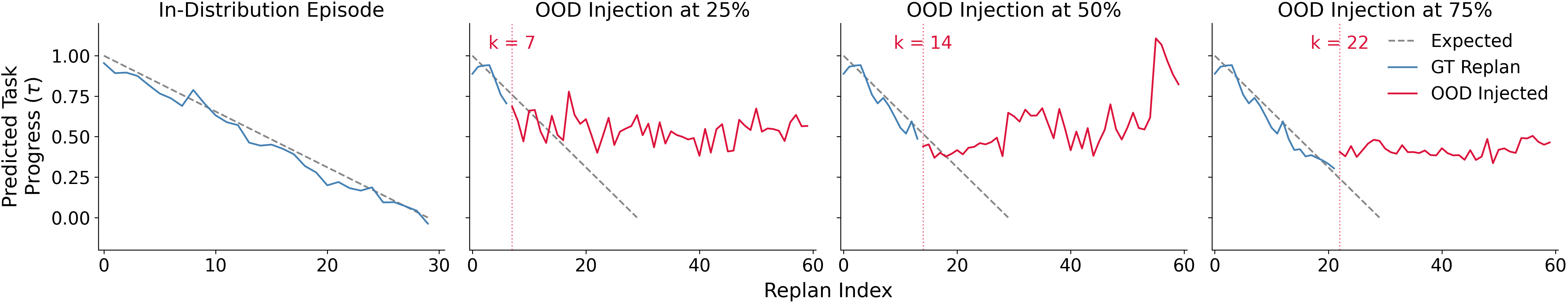}
    \caption{Qualitative OOD rollout under the \textbf{Blur} noise mode, with injections at $k = 0.25, 0.5, 0.75$.}
    \vspace{-8pt}
    \label{fig:qualitative_ood}
\end{figure}

\emph{Q1: Does the progress feature encode a strong indication of OOD?} Probing shows the progress feature loses its forward-progress signal the moment the robot enters an OOD state. Fig.~\ref{fig:qualitative_ood} shows the predicted progress qualitatively stagnating at all three injection points under perturbation. This deviation is easy to quantify and motivates a Gemma layer-0 OOD detector via Sec. ~\ref{eq:progress}.

\emph{Q2: Can the progress feature detect OOD states in new unseen tasks?}
We define unseen tasks as those outside the policy's training distribution. From the six tasks, we draw three at random, training all baselines and $\bm{V}_{\tau}$ on the ID and OOD data from those three, and evaluate on the remaining three across all $\binom{6}{3} = 20$ combinations. We compare against Mahalanobis~\citep{lee2018simple}, Variational Autoencoders (VAE), and SAFE-MLP/SAFE-LSTM~\citep{gu2025safe}. We evaluate at two granularities: \emph{per-replan} scores each model replan within a rollout, while \emph{per-episode} max-aggregates the replan scores to flag the rollout as success or failure, following SAFE~\citep{gu2025safe}.
\input{tables/3v3_unseen_seen}

On per-replan AUROC (Table~\ref{tab:3v3-seen-unseen}), $\bm{V}_{\tau}$ is the only detector whose per-replan performance does not degrade across the split ($0.796 \to 0.833$), consistent with its ID-only training; SAFE-MLP attains the strongest unseen score ($0.935$) but requires OOD supervision. Under per-episode aggregation, $\bm{V}_{\tau}$ achieves the best seen-task score ($0.910$) and is competitive on unseen tasks ($0.851$), trailing only SAFE-MLP ($0.917$).

\emph{Q3: Is the progress feature robust to new OOD states not seen in training?}
We test robustness to OOD types unseen during supervised training. Holding out two of the four noise modes at a time, we train the supervised baselines on the full six-task ID data plus the OOD data for the two seen modes and evaluate on the held-out modes, giving 3 splits. $\bm{V}_{\tau}$ is trained only on the six-task ID data and never sees any noise mode.

\input{tables/2v2_unseen_seen}

Table~\ref{tab:modes-seen-unseen-vprog-safe} reports AUROC pooled across modes when two are held out from supervised training. $\bm{V}_{\tau}$ wins both unseen columns (per-replan $\mathbf{0.871}$, per-episode $\mathbf{0.960}$), beating both supervised SAFE detectors despite never observing any OOD examples. Calibrating against expected progress thus yields a perturbation-agnostic failure signal that is both supervision-free and robust across OOD types, with the gap most pronounced at the episode level.

%% file: tables/3v3_unseen_seen.tex
\begin{table}[b]
\centering
\caption{Cross-task generalization: pooled AUROC of OOD detectors with tasks split into supervised-training (S) vs held-out (U). Per-replan and per-episode aggregations with mean $\pm$ std.}
\label{tab:3v3-seen-unseen}
\vspace{2pt}
\small
\setlength{\tabcolsep}{4pt}
\begin{tabular}{lcccc}
\toprule
& \multicolumn{2}{c}{per-replan} & \multicolumn{2}{c}{per-episode} \\
\cmidrule(lr){2-3} \cmidrule(lr){4-5}
Detector & S & U & S & U \\
\midrule
$V_{\tau}$ & $0.796\!\pm\!0.090$ & $0.833\!\pm\!0.058$ & $\mathbf{0.910}\!\pm\!0.036$ & $0.851\!\pm\!0.056$ \\
Mahalanobis & $\mathbf{0.986}\!\pm\!0.010$ & $0.859\!\pm\!0.040$ & $0.775\!\pm\!0.013$ & $0.719\!\pm\!0.016$ \\
VAE & $0.859\!\pm\!0.036$ & $0.796\!\pm\!0.056$ & $0.782\!\pm\!0.035$ & $0.778\!\pm\!0.034$ \\
SAFE-MLP & $0.917\!\pm\!0.025$ & $\mathbf{0.935}\!\pm\!0.013$ & $0.904\!\pm\!0.021$ & $\mathbf{0.917}\!\pm\!0.015$ \\
SAFE-LSTM & $0.827\!\pm\!0.041$ & $0.849\!\pm\!0.048$ & $0.836\!\pm\!0.035$ & $0.838\!\pm\!0.030$ \\
\bottomrule
\end{tabular}
\end{table}

%% file: tables/2v2_unseen_seen.tex
\begin{table}[b]
\centering
\vspace{-4pt}
\caption{Cross-OOD generalization on the modes experiment: pooled AUROC with mean $\pm$ std of $\bm{V}_{\tau}$ vs the two supervised SAFE detectors, with perturbation modes split into training (S) and held-out (U). $\bm{V}_{\tau}$ is mode-agnostic (never sees any OOD examples at training)}
\label{tab:modes-seen-unseen-vprog-safe}
\vspace{2pt}
\small
\setlength{\tabcolsep}{4pt}
\begin{tabular}{lcccc}
\toprule
& \multicolumn{2}{c}{per-replan} & \multicolumn{2}{c}{per-episode} \\
\cmidrule(lr){2-3} \cmidrule(lr){4-5}
Detector & S & U & S & U \\
\midrule
$V_{\tau}$ & -- & $\mathbf{0.871}\!\pm\!0.104$ & -- & $\mathbf{0.960}\!\pm\!0.013$ \\
SAFE-MLP & $0.839\!\pm\!0.039$ & $0.844\!\pm\!0.044$ & $0.859\!\pm\!0.048$ & $0.894\!\pm\!0.063$ \\
SAFE-LSTM & $0.637\!\pm\!0.149$ & $0.641\!\pm\!0.159$ & $0.835\!\pm\!0.061$ & $0.857\!\pm\!0.103$ \\
\bottomrule
\end{tabular}
\end{table}

%% file: sections/discussion.tex
Task progress is linearly readable from $\pi_{0.5}$'s activations, emerges from large-scale robotic pre-training, supports an unsupervised OOD detector, and varies under language counterfactuals. The feature is not steerable, exposing a gap between what the policy represents and what it acts on.

Our results are limited in scope. All quantitative experiments use a single model family ($\pi_{0.x}$) in simulation, leaving open whether the signal transfers across VLA architectures and to real-world deployment. Task progress is defined as normalized time in successful demonstrations, conflating elapsed time with semantic completion. The strong-decodability result under $\mathcal{T}_{\text{lang}}$ provides partial evidence against the pure-time interpretation, but a fully semantic definition requires completion labels. Similarly, the counterfactual labels used to train the augmented probe in Section~\ref{subsec:strong} are themselves proxies: we treat any swapped instruction as zero progress, since true semantic progress on a counterfactual task is not directly measurable without separate rollouts. The negative steerability result is also narrow, testing a single injection direction at a single layer with a fixed magnitude, and stronger or jointly-edited interventions may provide stronger ways forward for model steering.

Our work has several natural extensions, including evaluating progress decodability across VLA backbones, validating the OOD detector on real-robot rollouts, and probing for richer task-conditional signals beyond scalar progress. Our work suggests linearly readable semantic features can be a lightweight handle for monitoring deployed visuomotor policies, complementing the heavier interpretability toolkit increasingly applied to VLAs.

%% file: sections/expdesign.tex
\subsection{Models and Dataset}
\subsubsection{Dataset}

We run all experiments on VLABench~\citep{zhang2025vlabench}. The simulator provides a primitive dataset of 10 primary tasks:
\begin{enumerate}[nosep,leftmargin=*]
    \item Add Condiment: Pick up the target condiment and pour it over the dish.
    \item Select Fruit: Pick up the target fruit and place it on the plate.
    \item Select Drink: Pick up the target drink from the fridge.
    \item Insert Flower: Pick up the target flower and place it in the vase.
    \item Select Poker: Pick up the target playing card from the table.
    \item Select Mahjong: Pick up the target mahjong piece from the table.
    \item Select Book: Pull the target book from the shelf.
    \item Select Chemistry Tube: Touch the target tube in the chemical holder.
    \item Select Toy: Pick up the target toy and place it in the box.
    \item Select Painting: Press the button for the target painting.
\end{enumerate}

The dataset contains 5000 trajectories, 500 per task, with prompts and objects that vary across scenes to keep it diverse.

\subsubsection{Models}

We evaluate all our work on the following three models:
\begin{enumerate}[nosep,leftmargin=*]
    \item PaliGemma: a 400M SigLIP vision encoder followed by a 2B Gemma (18 transformer layers, $d = 2048$). It processes three camera images (base, left wrist, right wrist), which we remap to (wrist, front, side) for the agent and robot views.
    \item $\pi_0$ and $\pi_{0.5}$: each pairs a PaliGemma VLM backbone with a 300M Gemma for action generation.
\end{enumerate}

VLABench provides \textbf{fine-tuned} checkpoints of $\pi_{0}$ and $\pi_{0.5}$ on its primitive dataset of 5000 episodes, which we adopt directly. These checkpoints follow Physical Intelligence's training procedure on eight H100 GPUs. For the \textbf{base} $\pi_{0.5}$ model, we take Physical Intelligence's checkpoint and generate embeddings on the primitive dataset. For the \textbf{PaliGemma} checkpoint, we take Google's 3B model from HuggingFace and generate embeddings on the same dataset.

\subsubsection{Task Memorization}

We test whether the fine-tuned policy conditions its actions on the language instruction through two counterfactual interventions on the \textit{Select Fruit} task, where the instruction names a target fruit to place on the plate. The task's grasp progress measures object targeting: it advances only when the policy grasps the \emph{named} target (grasping any distractor leaves progress at zero), which gives a direct readout of which object the policy reaches for.

\textbf{Prompt swap.} We leave the scene unchanged and replace the target word in the instruction with a different in-scene object (e.g.\ ``Put the \text{pear}\dots'' $\rightarrow$ ``Put the \text{peach}\dots''), then execute the actions the policy produces under the altered prompt. In $10/10$ episodes the policy still reaches for the original target and ignores the swapped word entirely. It stays competent and directed, which isolates a failure to condition on language at the action head rather than an inability to act.

\textbf{Position swap.} We instead leave the instruction unchanged and swap the physical positions of the target and a random distractor before execution. The policy grasps the correctly relocated target in only $3/18$ episodes ($\sim$17\%) and reaches the original location in the rest. (Because progress conflates ``reached the wrong object'' with ``reached the right object but failed to grasp,'' this is a lower bound on language following.)

Both interventions show that the policy bases its actions on the memorized canonical scene layout rather than on the language instruction.

\subsection{Decodability \& Steerability versus Observability \& Controllability}

\citet{buurmeijer2026observing} define \textit{feature observability}: a feature is observable when some map exists that takes an embedding and outputs the corresponding feature.

\textbf{Weak Decodability.} In contrast, we measure decodability with linear maps alone, which requires the feature to be linearly recoverable from the representation and gives a more stringent criterion than observability. Because we aim to minimize risk, we also relax the equality in observability: rather than reproduce the feature exactly, the map need only recover it with low error, so we target high correlation instead of equality. Features may not be exactly recoverable from the representation, but as long as they are linearly decodable with low error, they are still useful for interpretation and control.

\citet{buurmeijer2026observing} also define \textit{feature controllability}: a feature is controllable when a map can propagate it through the activations by injecting a modified activation.

We move away from this definition because it only checks whether the feature can propagate; it does not (a) judge the quality of the feature given its inputs, nor (b) require the feature to cause changes in the model's behavior.

\textbf{Strong Decodability.} We call a feature \textit{strongly decodable} if perturbing its inputs drives its output to the ground truth those inputs imply. This is a stronger condition because it lets the interpreter define how the feature should depend on its inputs, for example requiring task progress to track the instruction and the trajectory time, which yields more natural conditions on the feature.

\textbf{Steerability.} A feature is steerable if perturbing it produces a new feature value that elicits the same response the model would give under the matching ground truth inputs. This establishes a form of causality between the feature and the model's behavior, since perturbing the feature reproduces the response we would see if that value were the true feature.

Together, these variations impose stronger and more varied conditions for controlling a feature by probing both its robustness and its causality, which are crucial for interpretability and control.

%as the ability to \textit{weakly decode} a feature from a representation using an observor. However, we choose to include \textit{strong decodability} as well, which allows us to measure the extent to which a feature can be perturbed by its inputs in a way that it outputs to a ground truth based on the feature. This is important as observing a feature requires the ability to control it, via perturbations. These two align with the concepts of \textit{feature observability} and \textit{feature steerability} defined by~\citep{buurmeijer2026observing}. Additionally, a core component of a feature should be a form of causality dictated by the transformations found through \textit{strong decodability}, leading to our definition of \textit{steerability} as the ability to perturb a feature to get the same respons from the model as if the underlying feature was the ground truth. This variation helps us to better understand the internal representations of the model and how they relate to the underlying features of the task, which is crucial for interpretability and control.

\subsection{Probe Setup and Training}

Algorithm \ref{alg:linear-probe} gives the procedure for training our linear probes.

\begin{algorithm}[tb]
\caption{Linear Progress Probe}
\label{alg:linear-probe}
\begin{algorithmic}[1]
\Require Probe dataset $\mathcal{D}_\text{probe}=\{(\bm{z}^i_t,\,\tau_t)\}$ at layer $i$,
  from $N$ trajectories $\xi$ with timestep $t$ and horizon $T$
\Require Epochs $E$, batch size $B$, learning rate $\eta$, validation fraction $\rho$
\Statex \textbf{Build progress labels (normalized time remaining, Eq.~\ref{eq:progress})}
\For{each trajectory $\xi$, each timestep $t$}
  \State $\tau_t \gets 1 - t/T \in [0,1]$
    \Comment{$1$ at start, $0$ near success}
\EndFor
\State Partition trajectories into $\mathcal{D}_{\text{val}}$ ($\lceil \rho N\rceil$ trajectories)
  and $\mathcal{D}_{\text{train}}$ \Comment{trajectory-level split, no leakage}
\Statex \textbf{Train probe} $\varphi(\bm{z}^i) = \bm{w}^\top \bm{z}^i + b$, parameters $\theta=\{\bm{w},b\}$, $\bm{w}\in\mathbb{R}^d$
\State Init $\theta$; optimizer $\gets \textsc{Adam}(\eta)$; $\text{best} \gets \infty$
\For{$\text{epoch} = 1 \ldots E$}
  \For{each minibatch $(\bm{z}_b, \tau_b)$ of size $B$ from a random permutation of $\mathcal{D}_{\text{train}}$}
    \State $\mathcal{L} \gets \frac{1}{|\bm{z}_b|}\sum \lVert \varphi(\bm{z}_b) - \tau_b \rVert_1$
      \Comment{$L_1$ loss}
    \State $\theta \gets \theta - \eta\,\nabla_\theta \mathcal{L}$
  \EndFor
  \State $m \gets R\big(\varphi;\,\mathcal{D}_{\text{val}}\big)$
    \Comment{validation MAE (risk under $L_1$)}
  \If{$m < \text{best}$} \State $\text{best} \gets m$;\ $\theta^\star \gets \theta$
    \Comment{keep best-on-val-MAE checkpoint}
  \EndIf
\EndFor
\State \Return $\varphi$ with $\theta^\star$; validation MAE and $R^2$
\end{algorithmic}
\end{algorithm}

We define the MLP Sweep experiment in Figure \ref{fig:observability} (right) through Algorithm \ref{alg:mlp-probe}. This capacity sweep trains probes of varying hidden widths and compares their performance on the true progress labels against shuffled labels (the control task). We ask whether added capacity improves performance on the true progress task and, if so, whether that improvement is significantly greater than any gain on the control task. A significant gap between the true and control tasks at high capacity would suggest that the representations hold nontrivial progress information that a sufficiently powerful probe can extract.

\begin{algorithm}[tb]
\caption{MLP Progress Probe (capacity sweep)}
\label{alg:mlp-probe}
\begin{algorithmic}[1]
\Require Embeddings and progress labels $(\bm{z}^i_t, \tau_t)$ as in Alg.~\ref{alg:linear-probe};
  hidden widths $\mathcal{H}$ (e.g.\ $\{0, 32, 64, 128, 256, 512, 1024\}$)
\Require Epochs $E$, batch size $B$, learning rate $\eta$, validation fraction $\rho$
\Statex \textbf{Width-$h$ probe (single hidden layer):}
  $\varphi_h(\bm{z}^i)=\bm{W}_2\,\mathrm{GELU}(\bm{W}_1\bm{z}^i+\bm{b}_1)+b_2$, with
  $\bm{W}_1\in\mathbb{R}^{h\times d}$, $\bm{W}_2\in\mathbb{R}^{1\times h}$;
  $h=0$ drops the hidden layer and recovers the linear probe $\varphi$ of Alg.~\ref{alg:linear-probe}
\For{each hidden width $h \in \mathcal{H}$}
  \For{label condition $c \in \{\text{true},\ \text{shuffled}\}$} \Comment{shuffled = control task}
    \State If $c=\text{shuffled}$: randomly permute labels $\{\tau_t\}$ across all embeddings
    \State Trajectory-level split into $\mathcal{D}_{\text{train}}, \mathcal{D}_{\text{val}}$ (fraction $\rho$, fixed seed)
    \State Init $\varphi_h$ (parameters $\theta$); optimizer $\gets \textsc{Adam}(\eta)$; $\text{best} \gets \infty$
    \For{$\text{epoch} = 1 \ldots E$}
      \For{each minibatch $(\bm{z}_b, \tau_b)$ of size $B$}
        \State $\theta \gets \theta - \eta\,\nabla_\theta\, R\big(\varphi_h;\,\{(\bm{z}_b, \tau_b)\}\big)$
          \Comment{$L_1$ loss}
      \EndFor
      \State $\text{best} \gets \min\!\big(\text{best},\ R(\varphi_h;\,\mathcal{D}_{\text{val}})\big)$
        \Comment{validation MAE}
    \EndFor
    \State Record best train/val MAE for $(h, c)$
  \EndFor
\EndFor
\State \Return per-width train/val MAE for true and shuffled probes
\end{algorithmic}
\end{algorithm}

For each probe, we use the hyperparameters in Table \ref{tab:linear-probe-hparams}. The MLP Sweep probes use the same settings, except that we also sweep the hidden widths as Algorithm \ref{alg:mlp-probe} describes.

\begin{table}[tb]
\centering
\caption{Linear probe training hyperparameters.}
\label{tab:linear-probe-hparams}
\begin{tabular}{ll}
\toprule
Hyperparameter & Default \\
\midrule
Epochs          & $300$ \\
Batch size      & $64$ \\
Learning rate   & $1\mathrm{e}{-3}$ (Adam) \\
Test fraction   & $0.2$ (episode-level split) \\
Seed            & $42$ \\
\bottomrule
\end{tabular}
\end{table}

\subsection{Contrastive Probe Training}
\label{app:contrastive}
A naive counterfactual-augmented probe does not yield strong decodability: because we assign every swapped activation the same constant target, $\varphi$ can fit it by shifting its absolute output level rather than learning the direction that distinguishes the two prompts, so its outputs under the original and swapped prompts barely separate. We instead train with a within-pair hinge that pairs each $\bm{z}_t$ with the swapped activation $\bm{z}'_t$ at the same step and separates them by a progress-dependent margin, $\big[\,m(t) + \varphi(\bm{z}_t) - \varphi(\bm{z}'_t)\,\big]_+^2$, where $m(t)$ grows linearly along the trajectory from $0.5$ at the start to $2.5$ at completion (the margin stays small early, when the original and swapped tasks look alike, and grows as the instructed task advances). This forces $\varphi$ onto the relative, instruction-conditioned direction.

The hinge constrains only the difference $\varphi(\bm{z}_t)-\varphi(\bm{z}'_t)$ and leaves the absolute level free, which saturates under a sigmoid readout and destroys the progress signal. We therefore add a BCE anchor on the absolute labels ($\lambda \approx 1.5$) so the same probe both separates prompts (large $\Delta$) and stays progress-readable ($R^2$); the anchor weight trades the two off directly. We also sharpen the original-prompt target, $\tilde\tau_t = \sigma((\tau_t - 0.5)/0.12)$, which improves progress calibration and raises the plateau $R^2$ from ${\sim}0.24$ to ${\sim}0.34$. The probes are linear ($\varphi$, as in Alg.~\ref{alg:linear-probe}); we train them over $5$ seeds with activations clipped to $[-10,10]$, and pair construction caches $\geq 5$ distinct target-swap shadow passes per timestep.

\subsection{Steerability Equations}

For our steerability experiment in Section \ref{subsec:steerable}, we define the function $\mathcal{T}_\text{boost}$, which advances the input vector to a target progress $\tau_\text{boost}$. Following Eq. \ref{eq:steer}, we define the steering function as follows:

\begin{equation*}
  \tilde{\bm{z}}^i(\bm{x})
  \;=\;\bm{z}^i(\bm{x})
       + \frac{\zeta(\mathcal{T}_\text{boost}(\bm{x}))-\zeta(\bm{x})}
              {\|\bm{w}\|^2}\,\bm{w},
\end{equation*}

\begin{equation*}
  \tilde{\bm{z}}^i(\bm{x})
  \;=\;\bm{z}^i(\bm{x})
       + \frac{\tau_\text{boost}-\tau}
              {\|\bm{w}\|^2}\,\bm{w}
\end{equation*}

where $\tau$ is the current progress at $\bm{x}$, and $\tau_\text{boost}$ is the boosted progress produced by $\mathcal{T}_\text{boost}$.

\subsection{OOD Modes}

Table \ref{tab:ood-hyperparameters} lists the hyperparameters we use for each OOD mode.
\begin{table}[tb]
\centering
\caption{Per-mode image-OOD perturbation hyperparameters for dataset
collection. $\sigma$ is the severity passed to
\texttt{inject\_image\_ood\_from\_demos.py}; we apply the perturbations to
each camera view.}
\label{tab:ood-hyperparameters}
\setlength{\tabcolsep}{6pt}
\renewcommand{\arraystretch}{1.1}
\begin{tabular}{lll}
\toprule
Parameter & Value & Units \\
\midrule
\multicolumn{3}{l}{\textit{Gaussian noise} \quad ($\sigma = 0.1$)} \\
Noise distribution & $\mathcal{N}(0,\ \sigma\!\cdot\!255)$ & gray levels \\
Noise std          & $25.5$                                & gray levels \\
\midrule
\multicolumn{3}{l}{\textit{Color (HSV shift)} \quad ($\sigma = 0.3$)} \\
Hue shift $\Delta h$  & $U(-54,\ +54)$       & OpenCV ($\pm108^\circ$) \\
Saturation multiplier & $U(0.7,\ 1.3)$       & ---                     \\
Value multiplier      & $U(0.7,\ 1.3)$       & ---                     \\
\midrule
\multicolumn{3}{l}{\textit{Gaussian blur} \quad ($\sigma = 0.3$)} \\
Kernel size         & $15$            & pixels \\
Gaussian $\sigma_x$ & $6.0$           & pixels \\
\midrule
\multicolumn{3}{l}{\textit{Occlusion} \quad ($\sigma = 0.2$)} \\
Patch size       & $44 \times 44$                              & pixels (at 224$\times$224) \\
Number of patches& 3                                           & per frame \\
Patch position   & $U(0, H{-}p_h)\!\times\!U(0, W{-}p_w)$      & pixels    \\
Fill value       & $0$ (black)                                 & uint8     \\
\bottomrule
\end{tabular}
\end{table}

%% file: sections/moreexp.tex
\subsection{Generalization to Unseen Tasks}

\begin{figure}[tb]
    \centering
    \includegraphics[width=\textwidth]{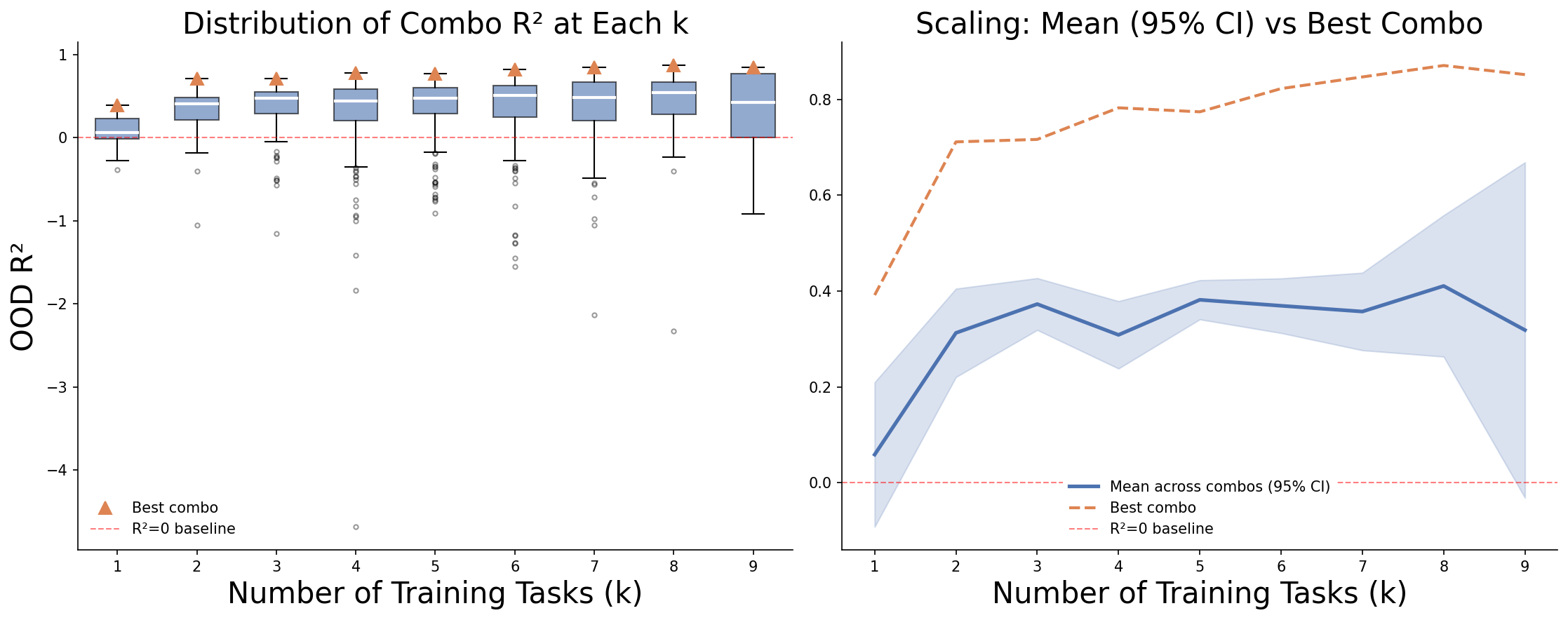}
    \caption{\textbf{The progress probe's generalization across tasks improves with the number of training tasks.} Out-of-distribution $R^2$ of progress probes evaluated on the unseen tasks as a function of the number of training tasks $k$. \textbf{Left:} distribution of OOD $R^2$ across all combinations of $k$ tasks (box plots), with the best combination (orange triangle) and the $R^2=0$ baseline (red dashed) marked. \textbf{Right:} mean across combinations (blue, $95\%$ CI) and the best combination (orange dashed) at each $k$. The best attainable $R^2$ rises from $\approx 0.39$ at $k=1$ to $\approx 0.85$ by $k=8$ before saturating, while the mean improves early and then plateaus with growing variance.}
    \label{fig:generalization-scaling}
\end{figure}

We measure how well a progress probe trained on $k$ tasks generalizes to the remaining unseen tasks, sweeping $k$ from $1$ to $9$ and scoring every combination of $k$ training tasks by its out-of-distribution (OOD) $R^2$ on those unseen tasks. Both training and evaluation use the full dataset of 5000 episodes (500 per task). Figure~\ref{fig:generalization-scaling} reports both the full distribution over combinations (left) and the mean and best combination at each $k$ (right). Generalization improves with more training tasks: the best combination climbs from $R^2 \approx 0.39$ at $k=1$ to $0.71$ at $k=2$ and up to $\approx 0.85$ by $k=8$, after which it saturates. The mean across combinations shows the same early gain (from $\approx 0.05$ at $k=1$ to $\approx 0.37$ at $k=3$) but then plateaus, and the spread widens at larger $k$, with a noticeably wider confidence interval at $k=9$. Most combinations sit comfortably above the $R^2=0$ baseline, yet the long negative tail at every $k$ shows that some task mixtures still transfer poorly. Adding training tasks therefore raises how well the progress feature generalizes across tasks, while the specific tasks we include remain a strong source of variance.

\subsection{Per-Task and Per-Mode OOD Results}

\input{tables/3v3_per_task_per_episode}
\input{tables/3v3_per_task_per_replan}

\paragraph{Per-task generalization (episode level).} Table~\ref{tab:3v3-per-task-per-episode} breaks down the pooled result of Table~\ref{tab:3v3-seen-unseen} by task; columns S/U give a task's AUROC when it was among the three seen vs.\ three unseen tasks of the split. Here $\bm{V}_{\tau}$'s only real competition is the supervised SAFE-MLP, and the two are closely matched: $\bm{V}_{\tau}$ takes the best seen score on four of six tasks (drink, fruit, mahjong, toy), and on unseen tasks it stays within $\sim$2 points of SAFE-MLP on add\_cond, fruit, and mahjong (winning mahjong outright, $0.899$), while SAFE-MLP leads clearly only on drink, flower, and toy. $\bm{V}_{\tau}$ achieves this with no OOD supervision, whereas SAFE-MLP trains on the seen tasks' failures; the remaining baselines trail both across every task.

\paragraph{Per-task generalization (replan level).} At the finer replan granularity (Table~\ref{tab:3v3-per-task-per-replan}) the ranking changes and $\bm{V}_{\tau}$ is no longer at the front. Mahalanobis is nearly perfect on \emph{seen} tasks ($\geq 0.976$ in every column) but falls off sharply on the unseen tasks (e.g.\ add\_cond $0.976\!\to\!0.761$), a memorization pattern rather than transferable detection; SAFE-MLP is the steadiest detector and best on every \emph{unseen} task ($0.943$--$0.978$). $\bm{V}_{\tau}$ sits in the middle of the pack here, strongest on drink ($0.901$ unseen) but weaker on the harder tasks, and only becomes competitive with the supervised SOTA once we aggregate replan scores to the episode level, echoing the results by mode below.

\input{tables/2v2_per_OOD_per_episode}
\input{tables/2v2_per_OOD_per_replan}

These tables decompose the pooled result of Table~\ref{tab:modes-seen-unseen-vprog-safe} by individual perturbation mode, at the episode level (Table~\ref{tab:modes-per-mode-per-episode}) and replan level (Table~\ref{tab:modes-per-mode-per-replan}). For each mode, column~S reports AUROC when that mode was \emph{seen} during supervised training and column~U when it was held out (\emph{unseen}); since $\bm{V}_{\tau}$ trains on ID data only, its columns report unseen rollouts in every case, whereas the supervised SAFE detectors expose a seen$\to$unseen gap that measures generalization across OOD modes.

\paragraph{Per-episode.} $\bm{V}_{\tau}$ attains the best unseen AUROC in \emph{every} mode (gaussian $0.963$, color $0.950$, blur $0.977$, occlusion $0.948$), exceeding the unseen score of both supervised SAFE detectors despite never observing a single OOD example. SAFE-MLP is the stronger baseline and transfers cleanly across the seen$\to$unseen split (often matching or exceeding its own seen score on the unseen modes, e.g.\ occlusion $0.823\!\to\!0.934$), yet its absolute unseen AUROC still trails $\bm{V}_{\tau}$ in every mode; SAFE-LSTM is weakest throughout.

\paragraph{Per-replan and the role of aggregation.} At the finer replan granularity the progress signal depends on the mode: $\bm{V}_{\tau}$ dominates gaussian ($0.938$) and blur ($0.994$) but trails SAFE-MLP on color ($0.750$ vs.\ $0.811$) and occlusion ($0.802$ vs.\ $0.858$). Crucially, aggregating the replan scores to the episode level by their maximum lifts $\bm{V}_{\tau}$ from the middle of the pack on color/occlusion to the best of the group on all four modes. Although its signal at the replan level is noisier under color shift and partial occlusion (perturbations that leave enough scene structure for the policy to keep emitting plausible actions that advance progress), it stays consistent enough over a trajectory to flag the failure once errors accumulate. This is precisely the regime, at the episode level, in which reliable failure detection matters, and it mirrors the pooled finding that $\bm{V}_{\tau}$'s advantage is largest there.

\subsection{Contrastive probing of the progress feature across layers}

\begin{figure}[t]
  \centering
  \includegraphics[width=\linewidth]{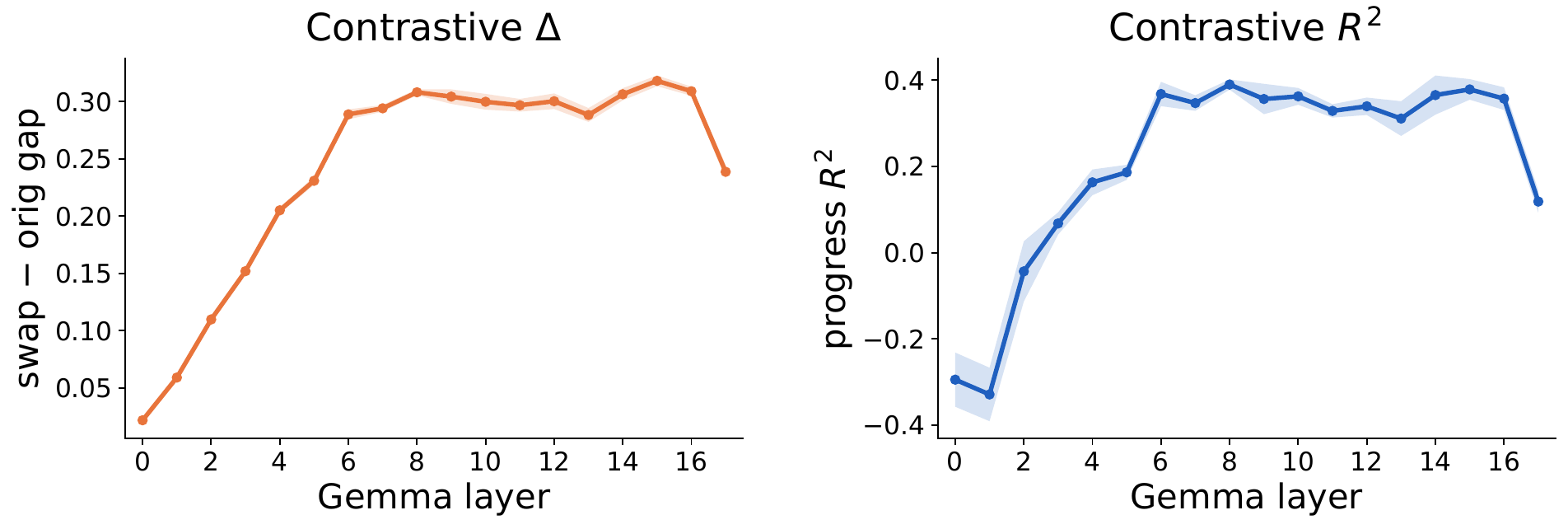}
  \caption{Language sensitivity $\Delta$ (left) and in-distribution progress
  $R^2$ (right) for the contrastive probe at each layer. Lines are the mean over
  $5$ seeds, bands $\pm1$ SEM.}
  \label{fig:contrastive-layer-sweep}
\end{figure}

We train the contrastive progress probe independently at every Gemma layer ($5$ seeds each) to choose the probing layer. For each layer we report two quantities (Fig.~\ref{fig:contrastive-layer-sweep}): the \emph{language sensitivity} $\Delta$, the mean gap over the trajectory between the probe's output under a prompt with the target swapped ($\mathcal{T}_\text{lang}$) and its output under the original instruction (large $\Delta$ = the probe responds to the instruction); and the in-distribution progress $R^2$ under the original instruction.

Both quantities are near zero in the earliest layers ($\Delta \approx 0$, $R^2 < 0$ for L0--L4), turn on together around L5--L6, and plateau across the middle and deep layers ($\Delta \approx 0.30$, $R^2 \approx 0.33$--$0.39$) before declining at the final layer (L17). The signal is therefore broadly available rather than localized to one site. We probe at \textbf{layer 10}: it sits on the plateau with nearly maximal $\Delta$ and $R^2$, away from the early layers that carry no signal and the final layer where both collapse.

\subsection{Data-Only Results}

\begin{figure}[tb]
    \centering
    \includegraphics[width=0.9\textwidth]{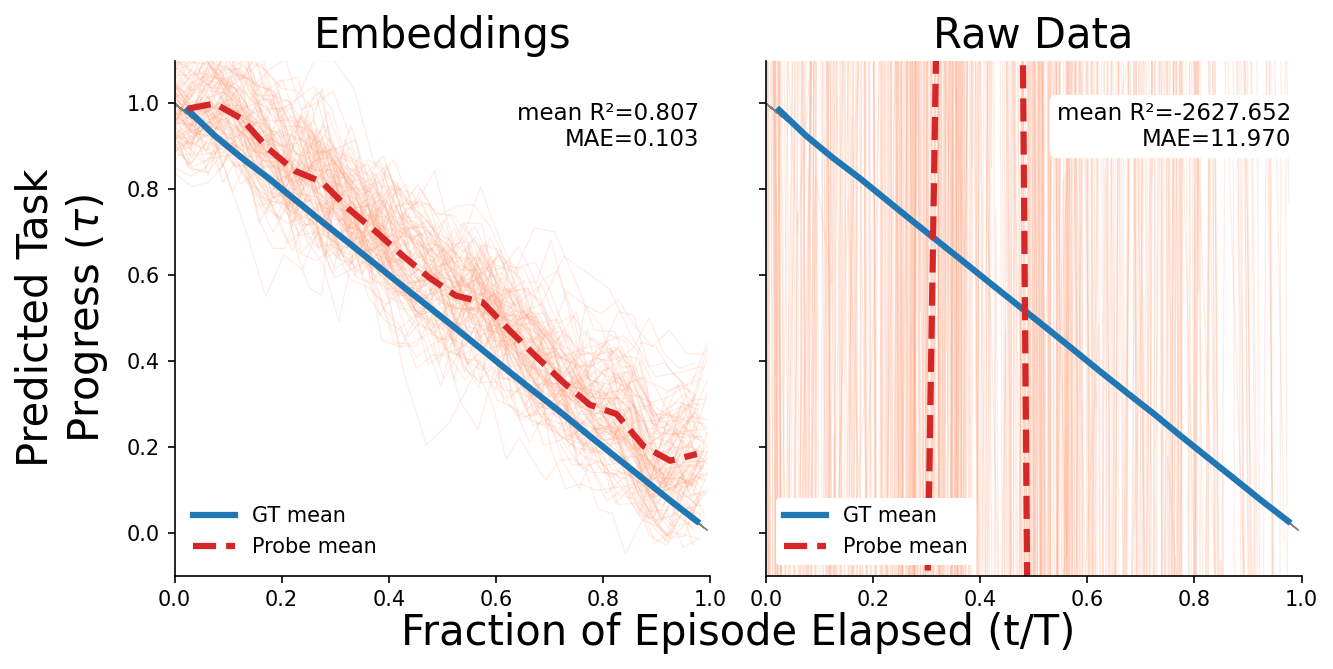}
    \caption{\textbf{Task progress is decodable from VLA embeddings but not from raw observations.} Predicted vs.\ ground truth task progress $\tau$ over normalized episode time $t/T$ for a probe trained on $\pi_{0.5}$ embeddings (left) and a matching probe trained on raw observations (right). Both probes are trained on 50 \emph{Add Condiment} episodes and evaluated on 100 out-of-distribution \emph{Add Condiment} episodes. The embedding probe tracks ground truth (mean $R^2 = 0.807$), while the probe on raw observations fails to generalize (mean $R^2 = -2627.7$, MAE $= 11.97$).}
    \label{fig:rawdata}
\end{figure}

To test whether task progress is a property of the VLA's learned representation rather than a signal already present in its inputs, we compare two probes: a linear probe on $\pi_{0.5}$ embeddings and, as a control, a linear probe trained directly on the raw observations. We train both on 50 episodes of \emph{Add Condiment} and evaluate them on 100 \emph{Add Condiment} episodes drawn from outside the training distribution. Figure~\ref{fig:rawdata} shows that the embedding probe recovers progress and tracks the ground truth closely (mean $R^2 = 0.807$, MAE $= 0.103$), whereas the probe on raw observations collapses out of distribution (mean $R^2 = -2627.7$, MAE $= 11.97$), its predictions diverging far outside the valid $[0,1]$ range. Task progress is therefore not linearly recoverable from raw observations; it is the VLA's representation that makes the feature decodable.

%% file: tables/3v3_per_task_per_episode.tex
% Per-task per-episode AUROC, seen vs unseen, mean $\pm$ std over 10 splits per cell
\begin{table}[tb]
\centering
\caption{Per-task AUROC of OOD detectors on the 3-seen vs 3-unseen task-split experiment, per-episode aggregation. For each task, columns S/U report AUROC when that task was in the seen / unseen set of the split (mean $\pm$ std over the 10 splits assigning that role). The six tasks are split across two row blocks for readability.}
\label{tab:3v3-per-task-per-episode}
\resizebox{\textwidth}{!}{%
\begin{tabular}{lcccccc}
\toprule
Detector & \multicolumn{2}{c}{$\text{add\_cond}$} & \multicolumn{2}{c}{$\text{flower}$} & \multicolumn{2}{c}{$\text{drink}$} \\
\cmidrule(lr){2-3} \cmidrule(lr){4-5} \cmidrule(lr){6-7}
  & S & U & S & U & S & U \\
\midrule
$V_{\tau}$ & $0.890\!\pm\!0.032$ & $0.866\!\pm\!0.036$ & $0.923\!\pm\!0.039$ & $0.879\!\pm\!0.041$ & $\mathbf{0.915}\!\pm\!0.027$ & $0.871\!\pm\!0.058$ \\
$\text{Mahalanobis}$ & $0.778\!\pm\!0.008$ & $0.718\!\pm\!0.013$ & $0.878\!\pm\!0.013$ & $0.713\!\pm\!0.018$ & $0.785\!\pm\!0.011$ & $0.749\!\pm\!0.006$ \\
$\text{VAE}$ & $0.722\!\pm\!0.020$ & $0.753\!\pm\!0.023$ & $0.842\!\pm\!0.041$ & $0.741\!\pm\!0.034$ & $0.816\!\pm\!0.014$ & $0.820\!\pm\!0.025$ \\
$\text{SAFE-MLP}$ & $\mathbf{0.891}\!\pm\!0.003$ & $\mathbf{0.889}\!\pm\!0.002$ & $\mathbf{0.951}\!\pm\!0.006$ & $\mathbf{0.952}\!\pm\!0.002$ & $0.893\!\pm\!0.025$ & $\mathbf{0.922}\!\pm\!0.002$ \\
$\text{SAFE-LSTM}$ & $0.782\!\pm\!0.018$ & $0.777\!\pm\!0.019$ & $0.854\!\pm\!0.047$ & $0.894\!\pm\!0.013$ & $0.804\!\pm\!0.052$ & $0.870\!\pm\!0.006$ \\
\bottomrule
\end{tabular}}

\vspace{2mm}

\resizebox{\textwidth}{!}{%
\begin{tabular}{lcccccc}
\toprule
Detector & \multicolumn{2}{c}{$\text{fruit}$} & \multicolumn{2}{c}{$\text{mahjong}$} & \multicolumn{2}{c}{$\text{toy}$} \\
\cmidrule(lr){2-3} \cmidrule(lr){4-5} \cmidrule(lr){6-7}
  & S & U & S & U & S & U \\
\midrule
$V_{\tau}$ & $\mathbf{0.923}\!\pm\!0.039$ & $0.877\!\pm\!0.043$ & $\mathbf{0.922}\!\pm\!0.039$ & $\mathbf{0.899}\!\pm\!0.043$ & $\mathbf{0.906}\!\pm\!0.041$ & $0.817\!\pm\!0.059$ \\
$\text{Mahalanobis}$ & $0.763\!\pm\!0.020$ & $0.745\!\pm\!0.007$ & $0.714\!\pm\!0.009$ & $0.719\!\pm\!0.009$ & $0.778\!\pm\!0.004$ & $0.722\!\pm\!0.013$ \\
$\text{VAE}$ & $0.775\!\pm\!0.045$ & $0.729\!\pm\!0.033$ & $0.801\!\pm\!0.065$ & $0.818\!\pm\!0.035$ & $0.743\!\pm\!0.044$ & $0.718\!\pm\!0.023$ \\
$\text{SAFE-MLP}$ & $0.891\!\pm\!0.015$ & $\mathbf{0.886}\!\pm\!0.006$ & $0.818\!\pm\!0.012$ & $0.894\!\pm\!0.005$ & $0.897\!\pm\!0.013$ & $\mathbf{0.905}\!\pm\!0.004$ \\
$\text{SAFE-LSTM}$ & $0.811\!\pm\!0.018$ & $0.839\!\pm\!0.015$ & $0.800\!\pm\!0.037$ & $0.895\!\pm\!0.004$ & $0.836\!\pm\!0.009$ & $0.828\!\pm\!0.015$ \\
\bottomrule
\end{tabular}}
\end{table}

%% file: tables/3v3_per_task_per_replan.tex
% Per-task per-replan AUROC, seen vs unseen, mean $\pm$ std over 10 splits per cell
\begin{table}[tb]
\centering
\caption{Per-task AUROC of OOD detectors on the 3-seen vs 3-unseen task-split experiment, per-replan aggregation. For each task, columns S/U report AUROC when that task was in the seen / unseen set of the split (mean $\pm$ std over the 10 splits assigning that role). The six tasks are split across two row blocks for readability.}
\label{tab:3v3-per-task-per-replan}
\resizebox{\textwidth}{!}{%
\begin{tabular}{lcccccc}
\toprule
Detector & \multicolumn{2}{c}{$\text{add\_cond}$} & \multicolumn{2}{c}{$\text{flower}$} & \multicolumn{2}{c}{$\text{drink}$} \\
\cmidrule(lr){2-3} \cmidrule(lr){4-5} \cmidrule(lr){6-7}
  & S & U & S & U & S & U \\
\midrule
$V_{\tau}$ & $0.849\!\pm\!0.081$ & $0.871\!\pm\!0.042$ & $0.794\!\pm\!0.078$ & $0.833\!\pm\!0.085$ & $0.850\!\pm\!0.097$ & $0.901\!\pm\!0.040$ \\
$\text{Mahalanobis}$ & $\mathbf{0.976}\!\pm\!0.003$ & $0.761\!\pm\!0.011$ & $\mathbf{0.999}\!\pm\!0.000$ & $0.910\!\pm\!0.039$ & $\mathbf{0.996}\!\pm\!0.003$ & $0.885\!\pm\!0.012$ \\
$\text{VAE}$ & $0.811\!\pm\!0.031$ & $0.754\!\pm\!0.032$ & $0.877\!\pm\!0.047$ & $0.759\!\pm\!0.083$ & $0.910\!\pm\!0.035$ & $0.923\!\pm\!0.035$ \\
$\text{SAFE-MLP}$ & $0.953\!\pm\!0.007$ & $\mathbf{0.943}\!\pm\!0.002$ & $0.961\!\pm\!0.011$ & $\mathbf{0.955}\!\pm\!0.012$ & $0.979\!\pm\!0.003$ & $\mathbf{0.978}\!\pm\!0.003$ \\
$\text{SAFE-LSTM}$ & $0.900\!\pm\!0.037$ & $0.884\!\pm\!0.054$ & $0.921\!\pm\!0.040$ & $0.933\!\pm\!0.033$ & $0.933\!\pm\!0.038$ & $\mathbf{0.978}\!\pm\!0.004$ \\
\bottomrule
\end{tabular}}

\vspace{2mm}

\resizebox{\textwidth}{!}{%
\begin{tabular}{lcccccc}
\toprule
Detector & \multicolumn{2}{c}{$\text{fruit}$} & \multicolumn{2}{c}{$\text{mahjong}$} & \multicolumn{2}{c}{$\text{toy}$} \\
\cmidrule(lr){2-3} \cmidrule(lr){4-5} \cmidrule(lr){6-7}
  & S & U & S & U & S & U \\
\midrule
$V_{\tau}$ & $0.788\!\pm\!0.098$ & $0.784\!\pm\!0.086$ & $0.716\!\pm\!0.105$ & $0.773\!\pm\!0.076$ & $0.735\!\pm\!0.080$ & $0.870\!\pm\!0.047$ \\
$\text{Mahalanobis}$ & $\mathbf{0.998}\!\pm\!0.000$ & $0.932\!\pm\!0.039$ & $\mathbf{0.999}\!\pm\!0.000$ & $0.885\!\pm\!0.021$ & $\mathbf{0.996}\!\pm\!0.000$ & $0.934\!\pm\!0.028$ \\
$\text{VAE}$ & $0.820\!\pm\!0.057$ & $0.729\!\pm\!0.044$ & $0.848\!\pm\!0.049$ & $0.809\!\pm\!0.030$ & $0.902\!\pm\!0.040$ & $0.846\!\pm\!0.051$ \\
$\text{SAFE-MLP}$ & $0.956\!\pm\!0.010$ & $\mathbf{0.945}\!\pm\!0.018$ & $0.974\!\pm\!0.020$ & $\mathbf{0.948}\!\pm\!0.008$ & $0.952\!\pm\!0.014$ & $\mathbf{0.943}\!\pm\!0.008$ \\
$\text{SAFE-LSTM}$ & $0.830\!\pm\!0.067$ & $0.842\!\pm\!0.056$ & $0.713\!\pm\!0.050$ & $0.849\!\pm\!0.060$ & $0.793\!\pm\!0.027$ & $0.794\!\pm\!0.068$ \\
\bottomrule
\end{tabular}}
\end{table}

%% file: tables/2v2_per_OOD_per_episode.tex
% Per-mode per-episode AUROC, train vs held-out modes, mean $\pm$ std over 3 combos per cell
\begin{table}[tb]
\centering
\caption{Per-OOD-mode AUROC on the cross-OOD generalization experiment (per-episode aggregation). For each perturbation mode, columns S/U report AUROC when that mode was in the supervised training set (seen) vs held out (unseen). Mean $\pm$ std over the 3 combos assigning each role. $\bm{V}_{\tau}$ is ID-only so its S/U columns reflect only unseen rollouts; the SAFE detectors are supervised so the S$-$U gap measures cross-OOD generalization.}
\label{tab:modes-per-mode-per-episode}
\resizebox{\textwidth}{!}{
\begin{tabular}{lcccccccc}
\toprule
Detector & \multicolumn{2}{c}{$\text{gaussian}$} & \multicolumn{2}{c}{$\text{color}$} & \multicolumn{2}{c}{$\text{blur}$} & \multicolumn{2}{c}{$\text{occlusion}$} \\
\cmidrule(lr){2-3} \cmidrule(lr){4-5} \cmidrule(lr){6-7} \cmidrule(lr){8-9}
  & S & U & S & U & S & U & S & U \\
\midrule
$V_{\tau}$ &-- & $\mathbf{0.963}\!\pm\!0.008$ & -- & $\mathbf{0.950}\!\pm\!0.006$ & -- & $\mathbf{0.977}\!\pm\!0.004$ & -- & $\mathbf{0.948}\!\pm\!0.007$ \\
$\text{SAFE-MLP}$ & $\mathbf{0.847}\!\pm\!0.054$ & $0.879\!\pm\!0.027$ & $\mathbf{0.885}\!\pm\!0.003$ & $0.854\!\pm\!0.095$ & $\mathbf{0.883}\!\pm\!0.022$ & $0.908\!\pm\!0.073$ & $\mathbf{0.823}\!\pm\!0.072$ & $0.934\!\pm\!0.036$ \\
$\text{SAFE-LSTM}$ & $0.800\!\pm\!0.081$ & $0.867\!\pm\!0.037$ & $0.874\!\pm\!0.006$ & $0.795\!\pm\!0.127$ & $0.848\!\pm\!0.065$ & $0.882\!\pm\!0.104$ & $0.819\!\pm\!0.070$ & $0.883\!\pm\!0.148$ \\
\bottomrule
\end{tabular}}
\end{table}

%% file: tables/2v2_per_OOD_per_replan.tex
% Per-mode per-replan AUROC, train vs held-out modes, mean $\pm$ std over 3 combos per cell
\begin{table}[tb]
\centering
\caption{Per-OOD-mode AUROC on the cross-OOD generalization experiment (per-replan aggregation). For each perturbation mode, columns S/U report AUROC when that mode was in the supervised training set (seen) vs held out (unseen). Mean $\pm$ std over the 3 combos assigning each role. $\bm{V}_{\tau}$ is ID-only so its S/U columns reflect only unseen rollouts; the SAFE detectors are supervised so the S$-$U gap measures cross-OOD generalization.}
\label{tab:modes-per-mode-per-replan}
\resizebox{\textwidth}{!}{
\begin{tabular}{lcccccccc}
\toprule
Detector & \multicolumn{2}{c}{$\text{gaussian}$} & \multicolumn{2}{c}{$\text{color}$} & \multicolumn{2}{c}{$\text{blur}$} & \multicolumn{2}{c}{$\text{occlusion}$} \\
\cmidrule(lr){2-3} \cmidrule(lr){4-5} \cmidrule(lr){6-7} \cmidrule(lr){8-9}
  & S & U & S & U & S & U & S & U \\
\midrule
$V_{\tau}$ &-- & $\mathbf{0.938}\!\pm\!0.039$ & -- & $0.750\!\pm\!0.028$ & -- & $\mathbf{0.994}\!\pm\!0.006$ & -- & $0.802\!\pm\!0.022$ \\
$\text{SAFE-MLP}$ & $\mathbf{0.816}\!\pm\!0.048$ & $0.862\!\pm\!0.013$ & $\mathbf{0.861}\!\pm\!0.008$ & $\mathbf{0.811}\!\pm\!0.050$ & $\mathbf{0.848}\!\pm\!0.025$ & $0.846\!\pm\!0.065$ & $\mathbf{0.833}\!\pm\!0.060$ & $\mathbf{0.858}\!\pm\!0.034$ \\
$\text{SAFE-LSTM}$ & $0.536\!\pm\!0.173$ & $0.733\!\pm\!0.046$ & $0.742\!\pm\!0.017$ & $0.519\!\pm\!0.158$ & $0.645\!\pm\!0.171$ & $0.650\!\pm\!0.193$ & $0.624\!\pm\!0.174$ & $0.662\!\pm\!0.194$ \\
\bottomrule
\end{tabular}}
\end{table}